\documentclass[10pt,journal,compsoc,onecolumn]{IEEEtran}

\usepackage{cite}         
\usepackage{graphicx}
\usepackage{booktabs}
\usepackage[accsupp]{axessibility}  
\usepackage{hyperref}
\usepackage{subcaption}
\usepackage{ragged2e}
\usepackage[normalem]{ulem}
\usepackage{amsmath}
\usepackage{amssymb}
\usepackage{color}
\usepackage{multirow}
\usepackage{float}
\usepackage{xcolor}
\usepackage[super]{nth}
\usepackage{bbm}
\usepackage{dsfont}
\definecolor{blue-violet}{rgb}{0.54, 0.17, 0.89}

\definecolor{blue-t}{HTML}{5364cc}

\newcommand{\ie}{\textit{i.e. }}

\usepackage{pifont}
\newcommand{\xmark}{\ding{55}}%

\usepackage{enumitem}
\usepackage[capitalize]{cleveref}
\crefname{section}{Sec.}{Secs.}
\Crefname{section}{Section}{Sections}
\Crefname{table}{Table}{Tables}
\crefname{table}{Tab.}{Tabs.}
\definecolor{lightgreen}{RGB}{129,194,234}
\usepackage{thmtools}
\declaretheorem[numberwithin=section]{theorem}

\declaretheorem[sibling=theorem]{proposition}

\usepackage{stfloats}  
\usepackage{cuted}     

\begin{document}{\twocolumn}

\title{Hyper-spherical Optimal Transport for Semantic Alignment in Text-to-3D End-to-end Generation}

\author{
        Zezeng~Li,
	Weimin~Wang,
        Yuming~Zhao,
        Wenhai~Li,
        Na~Lei,
        and~Xianfeng Gu
\IEEEcompsocitemizethanks{\IEEEcompsocthanksitem This research was supported by the National Key R$\&$D Program of China under Grant No. 2021YFA1003003;  the Natural Science Foundation of China under Grant No. T2225012, No. 12494554, and No. 62306059.

\IEEEcompsocthanksitem Z. Li, W. Wang, Y. Zhao, W. Li and N. Lei are with the School of Software, Dalian University of Technology, Dalian, 116620, China.

\IEEEcompsocthanksitem X. Gu is with the Department of Computer Science and Applied Mathematics, State University of New York at Stony Brook, USA.

\IEEEcompsocthanksitem Corresponding author: Na~Lei (Email: nalei@dlut.edu.cn) and Weimin Wang (Email: wangweimin@dlut.edu.cn).}}

\IEEEtitleabstractindextext{%
\begin{abstract}
\justifying
Recent CLIP-guided 3D generation methods have achieved promising results but struggle with generating faithful 3D shapes that conform with input text due to the gap between text and image embeddings. To this end, this paper proposes HOTS3D which makes the first attempt to effectively bridge this gap by aligning text features to the image features with spherical optimal transport~(SOT). However, in high-dimensional situations, solving the SOT remains a challenge.
To obtain the SOT map for high-dimensional features obtained from CLIP encoding of two modalities, we mathematically formulate and derive the solution based on Villani's theorem, which can directly align two hyper-sphere distributions without manifold exponential maps. Furthermore, we implement it by leveraging input convex neural networks~(ICNNs) for the optimal Kantorovich potential. With the optimally mapped features, a diffusion-based generator is utilized to decode them into 3D shapes. Extensive quantitative and qualitative comparisons with state-of-the-art methods demonstrate the superiority of HOTS3D for text-to-3D generation, especially in the consistency with text semantics. 
Codes are available on \href{https://github.com/cognaclee/HOTS3D}{\textcolor{lightgreen}{HOTS3D}}.
\end{abstract}

\vspace{-2mm}
}

\maketitle

\IEEEraisesectionheading{\section{Introduction}\label{sec:intro}}
\IEEEPARstart{T}{ext-to-3D} generation endeavors to create 3D assets that are coherent with the input texts, which has the potential to benefit a wide array of applications such as animations, games, architecture, and virtual reality.  Despite the extensive needs, the production of premium 3D content often remains a daunting task.  Traditional process of creating 3D assets requires a significant investment of time and effort, even when undertaken by professional designers~\cite{lei2023s}. This challenge has prompted the development of text-to-3D methods. Recently, with the development of large language models and large-scale datasets, deep learning based text-to-3D models~\cite{lee2022understanding,Wang_2022_CVPR,jain2022dreamfiled,poole2022dreamfusion,lin2023magic3d,cheng2023sdfusion,Chen_2023_ICCV,wang2023prolificdreamer,zhenyu2023text2tex,wang2023nerf,ibarrola2023collaborative} have made remarkable progress and can generate diverse shapes from text prompts. The existing text-to-3D methods are either CLIP-guided end-to-end methods or optimization-based methods.
The former trains a mapper from CLIP~\cite{radford2021learning} image embeddings to the shape embeddings of a 3D shape generator and switches to the CLIP text embedding as the input at test phase~\cite{chen2018text2shape,Mittal_2022_CVPR,Sanghi_2022_CVPR,li2023diffusion,jun2023shap,cheng2023sdfusion}. The latter iteratively optimizes the CLIP similarity~(SDS) loss between a text prompt and renders images of a 3D scene representation~\cite{jain2022dreamfiled,Wang_2022_CVPR,khalid2022clipmesh,lee2022understanding,xu2023dream3d} or the score distillation sampling loss~\cite{poole2022dreamfusion,lin2023magic3d,Chen_2023_ICCV,wang2023prolificdreamer} to achieve semantic alignment.

\begin{figure}[t]
    \vspace{-5mm}
  \centering
   \includegraphics[width=1.0\linewidth]{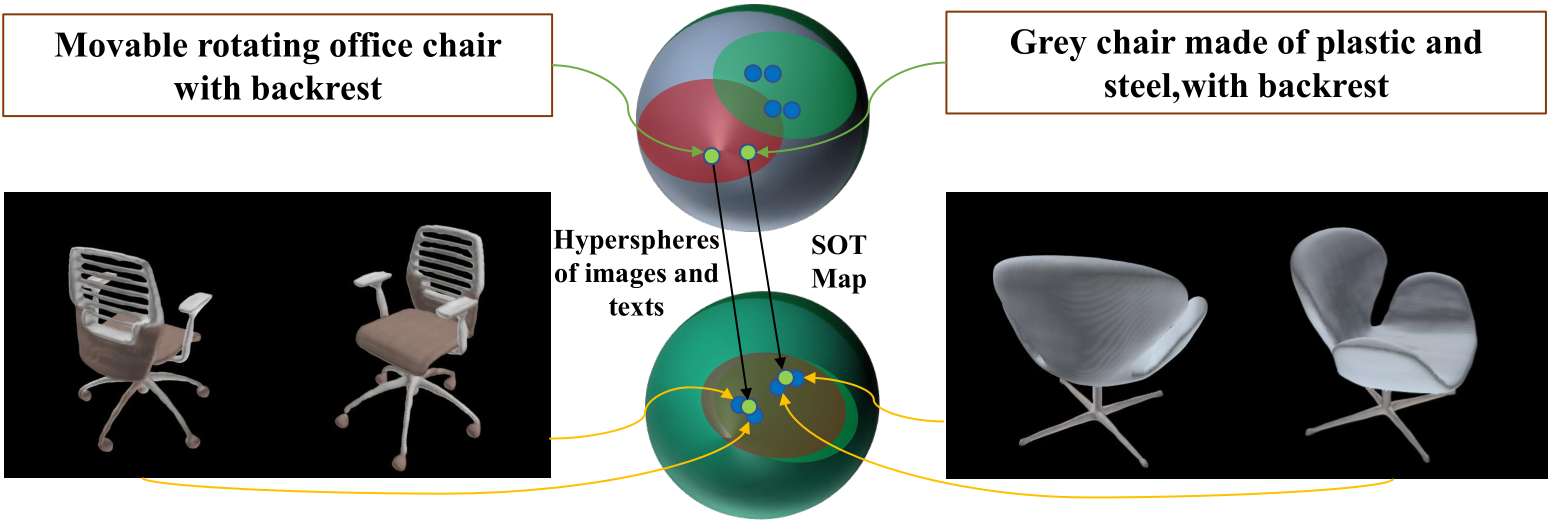}
      \caption{Semantic alignment with the proposed SOT map. \textcolor[RGB]{167, 95, 104}{Texts} and \textcolor[RGB]{57, 125, 86}{images} are embedded onto hyper-spheres by CLIP encoders. However, feature distributions~(ellipse regions) of two modalities may not be well aligned. SOT learns to optimally transport features to align them and thus enhances the semantic consistency of generated shapes. 
      }
   \label{fig:sot_illustration}
   \vspace{-3mm}
\end{figure}

CLIP-guided optimization-based methods typically exhibit superior text-shape consistency. However, they require re-optimizing model parameters for each text input, resulting in significant generation time costs. For instance, ProlificDreamer~\cite{wang2023prolificdreamer} takes over ten hours to generate a single 3D shape from a given text. In contrast, the end-to-end methods demonstrate notably faster generation speed with acceptable quality. 
Despite their efficiency, these methods often overlook the gaps between text and image features as illustrated in the upper sphere in Fig.~\ref{fig:sot_illustration}.   
This usually leads to a mismatch and semantic inconsistency of generated 3D shapes, especially in the testing stage, which treats the two equally.
Besides, as remarked in Dream3D~\cite{xu2023dream3d}, CLIP-guided methods tend to generate unconstrained “adversarial contents” that achieve high CLIP scores but with poor text consistency. To narrow the domain gap, Dream3D iteratively optimizes a text-to-image diffusion model, achieving more coherent results than other baselines. Nevertheless, as an optimization-based method, 
Dream3D inherently suffers from extended sampling duration, which is further exacerbated by an additional diffusion process.

To improve the semantic consistency of text-to-3D generation while maintaining efficiency, we consider introducing optimal transport~(OT) to align semantics and image features. Increasing OT-based methods are successfully applied in image generative models to align prior distribution for diverse generation~\cite{seguy2018large,chen2019gradual,An2020AEOT,liu2019wganqc,li2023dpm}. However, they typically operate in Euclidean space, while there exists the constraint of normalization on the hyper-sphere for CLIP-guided text and image features, \ie spherical optimal transport~(SOT). 
Furthermore, solutions for SOT are relatively less investigated. Especially for hyper-spheres, the implementation is challenging due to the high dimension. 

To this end, we formulate and derive the SOT map for high-dimensional semantic and image feature distribution alignment, leading to a novel hyper-sphere-OT-based end-to-end text-to-3D generative model, named~\textbf{HOTS3D}.
For SOT implementation, we approach SOT as a spherical Kantorovich dual problem and employ two input convex neural networks (ICNN)~\cite{amos2017input} to represent two convex potentials~\cite{villani2008optimal}, whose gradients are pivotal in solving SOT map. Once the solution is extrapolated, optimally mapped features are decoded into shapes. Our main contributions are: 

\begin{itemize}[leftmargin=*]
\item We are the first to formulate semantic alignment as a hypersphere OT problem for end-to-end text-to-3D generation. With the hyper-spherical constraint of CLIP features, we propose to optimally transport text features to image features on hyper-spheres to bridge the semantic gap. 
\item
We mathematically derive the solution of SOT on hyper-spheres for CLIP-extracted features. Furthermore, we design a whole pipeline for SOT learning and semantic alignment in text-to-3D generation.
\item 
We perform extensive experiments and analysis, which verify the superiority of the proposed \textbf{HOTS3D} in terms of semantic consistency and quality~(lower FID, P-FID, and LFD score) compared with state-of-the-art methods.
\end{itemize}
\vspace{-3mm}

\begin{figure*}
\vspace{-2mm}
  \centering
\includegraphics[height=0.55\linewidth,width=1.0\linewidth]{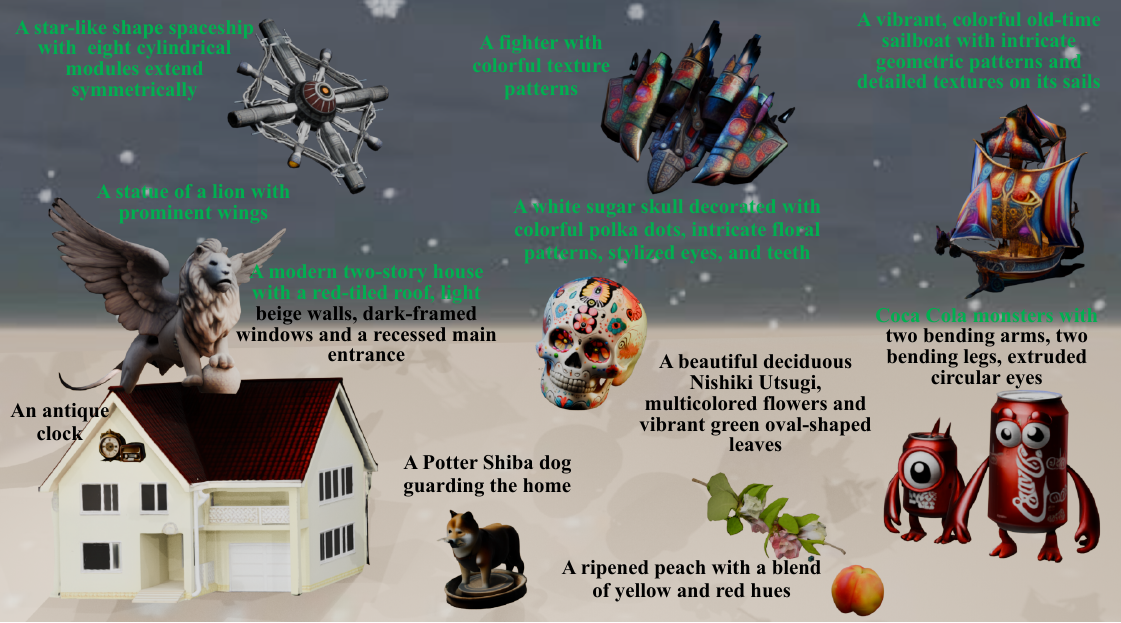}
   \caption{Monster Invasion, generated by HOTS3D}
   \label{fig:teaser}
   \vspace{-3mm}
\end{figure*}

\section{Related work}\label{sec:related}
\subsection{Text-to-3D content generation}
With the development of CLIP~\cite{radford2021learning} technology and the need for 3D content creation with simple text prompts, 
text-to-3D generation has attracted considerable attention~\cite{poole2022dreamfusion,jain2022dreamfiled,lin2023magic3d,zhenyu2023text2tex,long2023wonder3d,li2023sweetdreamer,liu2023unidream,yamada2024l3go}. 
Existing methods can be categorized CLIP-guided \textbf{end-to-end generative models} with pretrained~\cite{chen2018text2shape,Mittal_2022_CVPR,Sanghi_2022_CVPR,li2023diffusion,nam20223dldm,jun2023shap,cheng2023sdfusion,wei2023taps3d,sanghi2023clip,tang2023dreamgaussian,li2023instant3d,chen2023text,ren2024xcube,liu2023exim,zhao2024michelangelo,wu2024direct3d,zhang2024clay} 
or fine-tuned  CLIP encoders~\cite{liu2022towards,fu2022shapecrafter,tian2023shapescaffolder}, 
and CLIP-guided \textbf{optimization based methods}~\cite{jain2022dreamfiled,Wang_2022_CVPR,liu2022iss,khalid2022clipmesh,lee2022understanding,xu2023dream3d,poole2022dreamfusion,liu2023dreamstone,lin2023magic3d,Chen_2023_ICCV,wang2023prolificdreamer,qiu2023richdreamer,yi2024gaussiandreamer,chen2024vp3d,seolet,liang2024luciddreamer,qiu2023richdreamer,sun2024comogen,yang2024dreammesh}.
End-to-end generative models attempt to learn the distribution of the entire dataset, so that they can process different text inputs and generate various 3D assets without any fine-tuning. 
Optimization-based methods achieve bimodal alignment by iteratively optimizing the similarity between input text and the rendered images of the generated 3D assets. For those methods, it is necessary to finetune the model for each text, and the generation process is time-consuming. To facilitate the selection of baselines for experiments, we conducted a comprehensive investigation of the latest state-of-the-art baselines and summarized them in Tab.~\ref{tab:relatedwork}.

In this work, we focus on more efficient text-guided generation models, \ie end-to-end models. Among those existing methods, Shap·E~\cite{jun2023shap} is most relevant to this paper, which can generate diverse objects without relying on images as an intermediate representation and has gained popularity in the research community due to its impressive results. Recently, Diffusion-SDF~\cite{li2023diffusion} proposed an SDF autoencoder together with the voxelized Diffusion model to learn and generate representations for voxelized signed distance fields (SDFs) of 3D shapes. SDFusion~\cite{cheng2023sdfusion} enables 3D shape completion, reconstruction, and generation using a diffusion-based encoder-decoder. EXIM~\cite{liu2023exim} adopts a two-stage coarse-to-fine pipeline to generate shapes without time-consuming per-shape optimization. However, it requires training separate models for each category, such as chairs and tables, and the corresponding model of the target category needs to be called during generation. 
Direct3D~\cite{wu2024direct3d} focuses on image-to-3D generation, while text-to-3D generation is achieved by integrating existing text-to-image algorithms. Michelangelo~\cite{zhao2024michelangelo} aligns shape and clip text/image features through autoencoders to generate high-quality shapes, but there are still significant semantic inconsistencies in publicly available models.
These CLIP-guided text-to-3D content generation methods have yielded impressive results, but as remarked in Dream3d~\cite{xu2023dream3d,zhao2024michelangelo}, most of them overlooked the gap between the text and image embedding spaces. Consequently, semantic alignment is the key to filling this gap and improving text-shape consistency.

\begin{table}[t]
\footnotesize
\caption{{Recent works for text-to-3D generation~(as of 12/2024). E2E, Open, and Size denote end-to-end, open source code and models, and the size of test prompts. SIGA, SPN, OBJ, 3DF, and NA denote SIGGRAPH Asia, ShapeNet, Objaverse, 3D-FUTURE, and 'means unknown' respectively.}}
\vspace{-2mm}
\renewcommand{\arraystretch}{1.0}
\renewcommand{\tabcolsep}{1.1mm}
\centering
\begin{tabular}{l|c|c|c|c|c}
\hline
Methods&Venue&E2E&Open&3D dataset&Size\\ 
\hline
Kiss3DGen~\cite{lin2025kiss3dgen}&arXiv 2025&\checkmark &\checkmark& OBJ&100\\
TRELLIS~\cite{xiang2024structured}& arXiv 2024&\checkmark &\xmark& OBJ&NA\\
3DTopia-XL~\cite{chen20243dtopia}& arXiv 2024&\checkmark &\xmark& OBJ&NA\\
Ln3diff~\cite{lan2024ln3diff}& ECCV 2024&\checkmark &\checkmark& SPN \& OBJ&NA\\
Xcube~\cite{ren2024xcube}& CVPR 2024&\checkmark &\checkmark& SPN \& OBJ&30\\
EXIM~\cite{liu2023exim}&TOG 2023 &\checkmark &\checkmark& SPN \& 3DF&NA\\
Michelangelo~\cite{zhao2024michelangelo}&NIPS
2024 &\checkmark &\checkmark& SPN&2592\\
Diffusion-SDF~\cite{li2023diffusion}&CVPR 2023 &\checkmark &\checkmark& Text2Shape&NA\\
Shap$\cdot$E~\cite{jun2023shap}&arXiv 2023 &\checkmark &\checkmark& Point-e&NA\\
TAPS3D~\cite{wei2023taps3d}&CVPR 2023  &\checkmark &\checkmark&  SPN&1000\\
SDFusion~\cite{cheng2023sdfusion}&CVPR 2023 &\checkmark &\checkmark& Text2shape&NA\\
CLIP-Forge~\cite{Sanghi_2022_CVPR}&CVPR 2022 &\checkmark &\checkmark& SPN&234\\
\hline
Hunyuan3D~\cite{yang2024hunyuan3d}&arXiv 2024&\texttimes &\checkmark& OBJ&70\\
DreamMesh~\cite{yang2024dreammesh}&ECCV 2024&\texttimes &\xmark& T$^3$Bench&50\\
DiverseDream~\cite{DiverseDream}&ECCV 2024&\texttimes &\checkmark& None&70\\
COMOGen~\cite{sun2024comogen}&arXiv 2024&\texttimes &\xmark& OBJ&20\\
CLAY~\cite{zhang2024clay}& TOG 2024 &\texttimes &\xmark&  SPN \& OBJ&50\\
RichDreamer~\cite{qiu2023richdreamer}&CVPR 2024& \texttimes &	\checkmark & OBJ & 113\\
LucidDreamer~\cite{liang2024luciddreamer}&CVPR 2024 &\texttimes &\checkmark& Point-e&28\\
VP3D~\cite{chen2024vp3d}&CVPR 2024 &\texttimes &\xmark& T$^3$Bench&300\\
GaussDreamer~\cite{yi2024gaussiandreamer}&CVPR 2024 &\texttimes &\checkmark& Point-e&NA\\
ProlificDreamer~\cite{wang2023prolificdreamer}&NIPS
2024 &\texttimes &\checkmark& None&100\\
3DFuse~\cite{seolet}&ICLR
2024 &\texttimes &\checkmark& Co3D&42\\
Magic3D~\cite{lin2023magic3d}&CVPR 2023 &\texttimes &\xmark& None&397\\
Dreamfusion~\cite{poole2022dreamfusion}& ICLR 2023 &\texttimes &\xmark& None&153\\
PCLIPNeRF~\cite{lee2022understanding}&arXiv 2022&\texttimes &\checkmark& None&153\\
CLIP-Mesh~\cite{khalid2022clipmesh}&SIGA 2022 &\texttimes &\checkmark& None&NA\\
\hline
\end{tabular}
\label{tab:relatedwork}
\vspace{-3mm}
\end{table}	

\subsection{Semantic alignment}
Most text-to-3D generation methods are proposed based on the assumption that image features and text features correspond in the cross-modal embedding space, which is usually extracted by CLIP. Due to the issue of inconsistent text shapes in generation models, more and more researchers are attempting to align semantic and image features further to improve text-shape consistency. For example, Li et al.~\cite{li2021align} introduced a contrastive loss to align the image and text representations, making it easier for the multimodal encoder to perform cross-modal learning. Cheng et al.~\cite{cheng2021deep} designed a semantic alignment module to fully explore the latent correspondence between images and text. Guo et al.~\cite{guo2022tm2t} modeled the distribution over discrete motion tokens, which enables non-deterministic production of variable-length pose sequences for input texts. 3D-VisTA~\cite{zhu20233d} utilized a pre-trained Transformer for point cloud and text alignment, eliminating the need for auxiliary losses and optimization tricks in question-answering tasks. Nie~\cite{nie2023image} mined the semantic consistency by compacting each 3D shape and its nearest neighbors to enhance semantic alignment for unlabeled 3D model domains. Cao~\cite{cao2024coda} developed a cross-modal alignment
module to align feature spaces between 3D point clouds and image/text modalities. For consistent generation, Zhao~\cite{zhao2024michelangelo} proposed using one network to encode the 3D shapes into the shape latent space aligned to the image and text, and using another network to learn a probabilistic map from the image or text space to the latent shape space. Liu et al.~\cite{liu2022iss,liu2022towards,liu2023dreamstone} finetuned the text encoder to improve the text-shape consistency. Xu et al.~\cite{xu2023dream3d} bridged the text and image modalities with a diffusion model. These methods require iterative optimization of models with cosine similarity, or complex and time-consuming diffusion models, which motivates us to explore more effective alignment methods.

\subsection{Euclidean and spherical optimal transport} 
The OT map transmits one distribution to another in the most cost-effective manner, which has been widely applied for \textbf{distribution alignment} in Euclidean space~\cite{dominitz2009texture,seguy2018large,chen2019gradual,An2020AEOT,liu2019wganqc,li2022weakly,li2022real,li2023ot,li2023dpm,gu2022keypoint,Fan2021,Gazdieva2022}. Recently, several works have shown that using a prior on the hyper-sphere can improve the results~\cite{xu2018spherical,davidson2018hyperspherical,Gu_2020_CVPR,bonet2022spherical}. 
Contrary to Euclidean geometry, both the metric and the transportation costs under the spheres differ, leading to the traditional OT algorithms being ineffectual for spheres. 
Consequently, OT solutions for spheres or hyper-spheres have attracted community attention. Cui et al. \cite{cui2019spherical} provided a \textbf{2D SOT} algorithm from a geometric view to adjust the density of triangular facets for mesh parameterization. Later, Hamfeldt et al. \cite{hamfeldt2021convergent,hamfeldt2022convergence} proposed a PDE approach to numerically solving the 2D SOT map and constraining the solution gradient to obtain a stable scheme. Cohen et al.~\cite{cohen2021riemannian} designed a flow model based on the Riemannian exponential map to fit the 2D SOT. However, due to the challenge of calculating Jacobian determinants~(JD) in high dimensions, stochastic trace estimates need to be used to approximate JD for distribution transformation. Bonet et al.~\cite{bonet2022spherical} defined spherical sliced Wasserstein~(SSW) distances on Riemannian manifolds and utilized the Flow model~\cite{rezende2020normalizing} to fit the 2D SOT by minimizing the SSW loss. In high-dimensional cases, SSW was taken as a regularization term to guide the decoder to output a feature distribution aligning with the prior distribution, but the map fitted by the decoder is not an exact SOT map. For the \textbf{hyper-sphere OT}, 
Rezende et al.~\cite{rezende2021implicit} designed a Hessian matrix to reformulate the Kantorovich potential and gave a threshold criterion of the matrix for the hyper-sphere. Nonetheless, due to the difficulty in calculating the JD, only 2D SOT validation experiments were provided. As remarked in \cite{rezende2021implicit}, solving the SOT problem in high-dimensional settings based on exact likelihoods is still challenging. Moreover, SOT algorithms like Ref~\cite{rezende2021implicit, cohen2021riemannian} used Riemannian exponential maps to switch Euclidean data to hyper-spheres and then calculated the OT map in a Euclidean way, which inevitably leads to distortion of manifolds and accuracy decline of results. 

\begin{figure*}[t]
  \centering
  \vspace{-3mm}
   \includegraphics[width=0.9\linewidth]{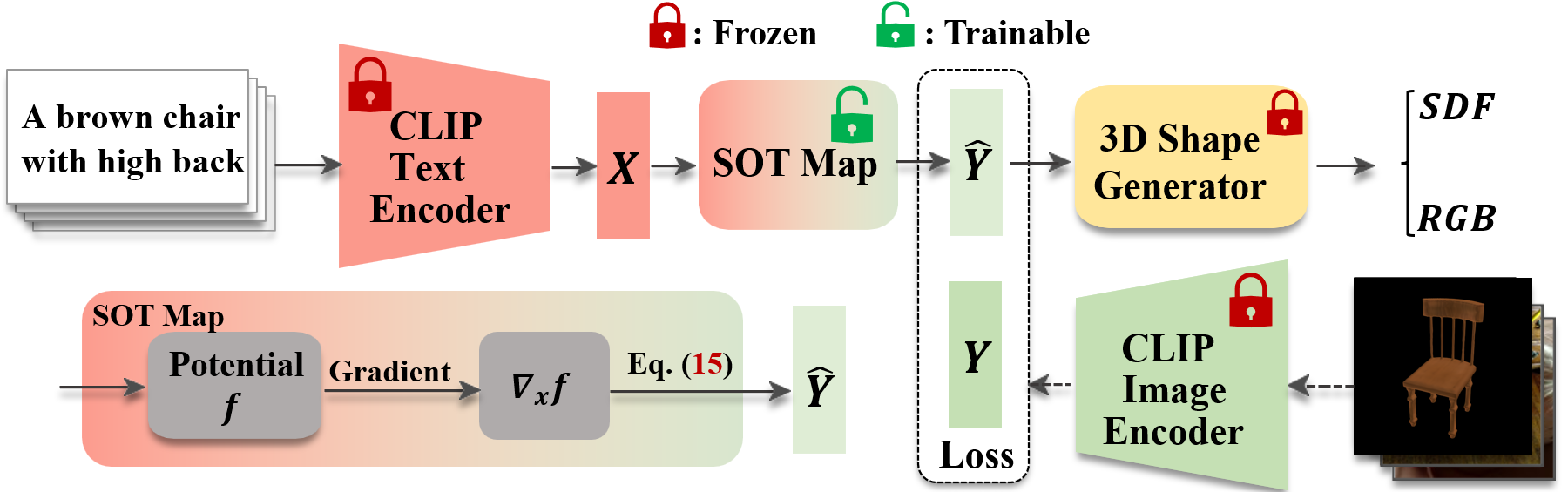}
   \caption{Overview of the proposed text-to-3D framework with the SOT map defined in Eq.~\eqref{eq:SOTMap}. The dotted line indicates the process only in the Potential function training phase to solve the SOT map. The Loss is calculated by Eq.~\eqref{eq:obj_f} and Eq.~\eqref{eq:obj_g}.}
   \label{fig:framework}
   \vspace{-3mm}
\end{figure*}

\section{Method}\label{sec:method}
\begin{figure*}[t]
  \centering
  \vspace{-3mm}
   \includegraphics[width=0.9\linewidth]{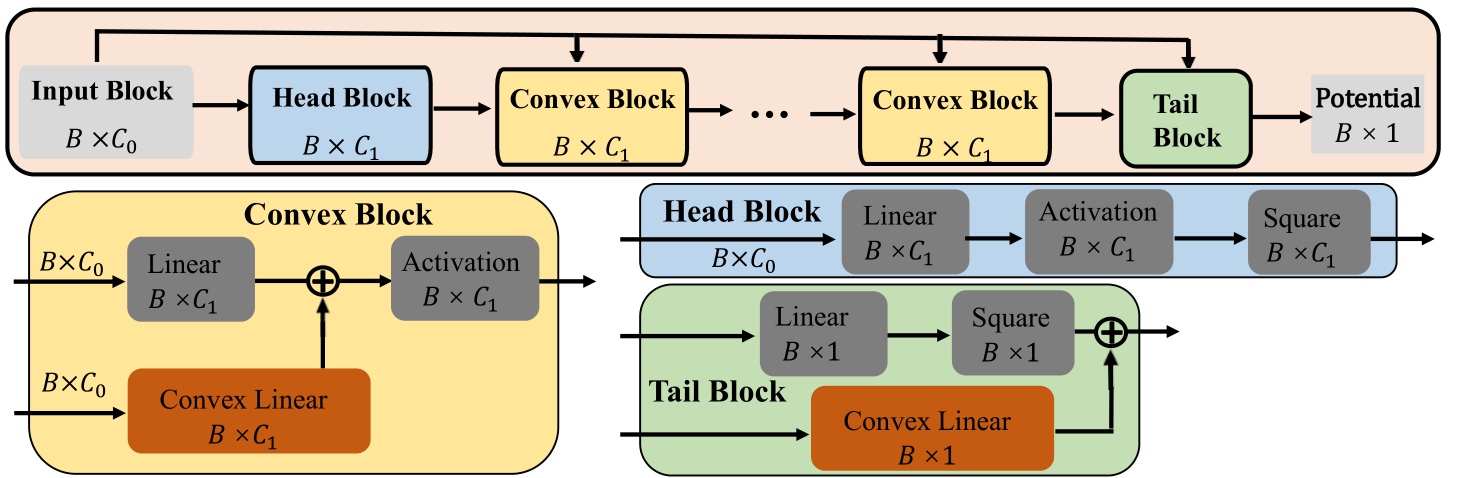}
   \caption{Network architecture of potential functions $f$ and $g$. $B$ denotes the batch size. $C_0$ and $C_1$ are the dimensions of the input features and hidden layer features, respectively. $\oplus$ is the addition in mathematics.}
   \label{fig:network}
   \vspace{-3mm}
\end{figure*}

Our objective is to generate 3D content that aligns with the given input text prompt. As illustrated in \cref{fig:framework}, our framework for text-guided 3D synthesis comprises three stages. Firstly, we encode the input text prompt onto the hyper-sphere with a pre-trained CLIP text encoder, obtaining text features. Secondly, the SOT map in Eq.\eqref{eq:SOTMap} is induced by the gradient of a convex function (see~\cref{fig:network}) that is trained via minimax optimization, and then transfers output text features to the image feature space. In the third stage, a generator conditioned on the output of the SOT Map was utilized to generate 3D shapes. The SOT map is a plug-and-play tool for aligning spherical distributions. During the training phase, we only need to optimize the parameters of the SOT map and other modules remain frozen, significantly reducing the training difficulty. With the SOT map for semantic alignment, our HOTS3D can bypass iterative
optimization during the testing phase, resulting in stronger generalization capability and semantic consistency.

\noindent\textbf{Observations and Motivations.} 
Though CLIP-guided E2E methods can generate diverse results, they are short in producing precise and detailed 3D structures that accurately match the text, as depicted in \cref{fig:teaser} and \cref{fig:text23D}. Its cause lies in the gaps between text and image features, especially in the testing stage, treating the two equally, directly leading to a mismatch issue. To narrow this gap, CLIP-guided optimization methods attempt to align the semantics of instance text through iteration, resulting in significant time costs. Thus, from the perspective of \textbf{distribution alignment}, we propose using SOT Map to transfer entire text features into the image space. For solutions of SOT, existing works are either not applicable to hyper-sphere, or cause distortion, which encourages us to design a more effective method. 

Compared with existing distribution alignment methods (e.g., GAN and VAE-based), the method proposed in this paper has the following advantages and theoretically essential differences: 1)~Space: our alignment is in hypersphere space, while most others are in Euclidean space. 2)~Align approach: ours fits a convex potential function whose gradient induces the transport map, while others fit the map with neural nets. Thus, they cannot represent discontinuous target distributions, leading to mode mixture~(mixed generation) results across boundaries. Conversely, ours can align distributions with discontinuous support while maintaining clear boundaries, as shown in \cref{fig:Ablation}.

\subsection{Problem formulation and preliminary}
\label{subsec:Setup}
Given text feature set $\boldsymbol{X} \subset \mathbb{S}^{d}$, image feature $\boldsymbol{Y} \subset \mathbb{S}^{d}$, and their respective prescribed probability measure $\mu_S$ and $\nu_S$, our goal is to transfer text features to the image feature space to achieve alignment of two modalities, thereby providing more accurate features encoding for generations. In the following, we introduce relevant OT theories to provide a preliminary for subsequent method explanations.

\noindent\textbf{Optimal Transport~(OT).} Let $\mu$ and $\nu$ be two probability distributions on $\mathbb{R}^d$. The \emph{Monge's optimal transport problem} is to transport the probability mass under $\mu$ to $\nu$ with the minimum total transportation cost, \ie
\begin{equation}
\underset{T: T_{\#}\mu=\nu} {\text{min}}\;\; \;\; \mathbb{E}_{\boldsymbol{X}\sim \mu}\ c(\boldsymbol{X},T(\boldsymbol{X})). \;
\label{eq:monge}
\end{equation}
$T$ achieving the minimum total cost in Eq.~\eqref{eq:monge} is called OT map. $T_{\#} \mu$ is the push-forward of $\mu$ under $T$. Subsequently, by combining the joint distribution,  Kantorovich relaxed the original \emph{Monge's OT problem} and provided a dual formulation \emph{Kantorovich duality problem} \cite{villani2008optimal}, 
\begin{align}
	W_d(\mu,\nu) =\; \sup_{(f,g)\in \Phi_c} \mathbb{E}_{\mu}[f(\boldsymbol{X})]+\mathbb{E}_{\nu}[g(\boldsymbol{Y})],
    \label{eq:dual_form}
\end{align}
where $\Phi_c$ denotes the constrained space of Kantorovich potentials, defined as $\Phi_c \triangleq \bigl\{(f,g)\in L^1(\mu)\times L^1(\nu): ~f(\boldsymbol{x})+g(\boldsymbol{y})\leq c(\boldsymbol{x},\boldsymbol{y}), \quad \forall (\boldsymbol{x},\boldsymbol{y}) ~d\mu \otimes d\nu~\text{a.e.}\bigr\}$. $c(\boldsymbol{x},\boldsymbol{y})$ is the cost function. The optimal transport map can be obtained from $f$ and $g$,  using the first-order optimality conditions of Fenchel-Rockafellar's duality theorem~\cite{seguy2018large}, or by training a generator through adversarial optimization~\cite{li2022weakly}. $f$ and $g$ play a symmetric role in Eq.~\eqref{eq:dual_form}, and either of them can be replaced by the other's $c$-transform. The $c$-transform of $g$ is defined by $g^c(\boldsymbol{x})=\inf_{{\boldsymbol{y}}}(c(\boldsymbol{x}, \boldsymbol{y})-g(\boldsymbol{y}))$. Then, as it is remarked in \cite[Theorem 5.26]{villani2009optimal} and \cite[Definition 3.3 $\&$ 3.6]{lei2019geometric,lei2020geometric}, the Kantorovich problem can be reformulated as the following dual problem:
\begin{align}\label{eq:dual_c_transform_form}
\hspace{-2pt}W_d(\mu,\nu) \!= \!\!\sup_{g \in L^1(\nu)} \mathbb{E}_{\mu}[g^c(\boldsymbol{X})]\!+\!\mathbb{E}_{\nu}[g(\boldsymbol{Y})],
\end{align}

\noindent\textbf{Spherical Optimal Transport~(SOT).} Given two prescribed probability measure $\mu_S$ and $\nu_S$ on $\mathbb{S}^{d}$, we seek a spherical map $T:\mathbb{S}^{d}\to\mathbb{S}^{d}$ such that
\begin{equation}
T = \min\limits_{T_\# \mu_S = \nu_S} \mathbb{E}_{X\sim \mu_S}\  c(\boldsymbol{X},T(\boldsymbol{X})).
\label{eq:SOT}
\end{equation}
Here $c(\boldsymbol{x},\boldsymbol{y})$ is the cost of transporting a unit of mass from $\boldsymbol{x}$ to $\boldsymbol{y}$. $\langle\cdot,\cdot\rangle$ denotes the vector inner product. $T_\# \mu_S = \nu_S$  indicates
$\int_{\boldsymbol{X}} \mu_S(\boldsymbol{x})\ d\boldsymbol{x} = \int_{T(\boldsymbol{X})} \nu_S(\boldsymbol{y})\ d\boldsymbol{y}$
for every measurable $\boldsymbol{X} \subset \mathbb{S}^{d}$, $\boldsymbol{Y} \subset \mathbb{S}^{d}$. Similarly, the spherical Kantorovich dual problem has the same form as Eq.~\eqref{eq:dual_form}.

\subsection{Derivation of SOT solution on hyper-sphere}
\label{sec:SOTmap}
Given the logarithmic transportation cost $c(\boldsymbol{x},\boldsymbol{y}) = \log(2- \langle \boldsymbol{x},\boldsymbol{y}\rangle )$ which is the widely used on spheres, potential  pair $(f,g) \in \Phi_c$ satisfy
\begin{equation}\label{constrain}
f(\boldsymbol{x})+g(\boldsymbol{y}) \leq \log(2- \langle \boldsymbol{x},\boldsymbol{y}\rangle )\,.
\end{equation}
To directly calculate the OT map on hyper-spheres, we reformulate dual problem in Eq.~\eqref{eq:dual_c_transform_form} into the following form involving only a single convex function $g$: 
\begin{align}\label{eq:dual_convex_form}
\hspace{-2pt}W_d \!= \!\!\sup_{g \in \mathtt{CVX}(\nu_S)} \mathbb{E}_{\mu_S}[g^c(\boldsymbol{X})]\!+\!\mathbb{E}_{\nu_S}[g(\boldsymbol{Y})]\,.
\end{align}
Here $\mathtt{CVX}(\nu)$ denotes the set of all convex functions in $L^1(\nu)$, $g^c(\boldsymbol{x})=\inf_{\boldsymbol{y}} (\log(2- \langle \boldsymbol{x},\boldsymbol{y}\rangle) - g(\boldsymbol{y}))$ is the conjugate c-transform of $g(\cdot)$.  According to \cite[Theorem 5.26]{villani2009optimal}, any optimal pair potential $(g,g^c)$ in Eq.~\eqref{eq:dual_convex_form} induce an OT map pushing forward $\nu_S$ onto $\mu_S$, and $(f,f^c)$ induce an OT map pushing forward $\mu_S$ onto $\nu_S$. Specifically, if $c(\boldsymbol{x}, \boldsymbol{y}) = \frac{1}{2}||\boldsymbol{x}-\boldsymbol{y}||^2$, the OT map $T(\boldsymbol{x})=\boldsymbol{x}-\nabla f(\boldsymbol{x})$. When we shift our attention to hyper-spheres, this solution no longer applies. To provide a functional relationship between the OT map and Kontarovich's potential on hyper-spheres, we introduce some theorems below.

\begin{theorem}
\label{thm:kp_optim_relation}
(\cite[Villani]{villani2009optimal}). Given $\mu$ and $\nu$ on a compact convex domain $\Omega \subset \mathbb{R}^{d}$, there exists an OT plan  for the cost $c(\boldsymbol{x}, \boldsymbol{y}) = h(\boldsymbol{x}-\boldsymbol{y})$, with $h$ strictly convex. It is unique and of the form $(id, T_{\#})\mu$ (id: identity map), provided that $\mu$ is absolutely continuous and $\partial\Omega$ is negligible. Moreover, there exists a Kantorovich's potential $\varphi$, and the OT map $T$ can be represented as follows: 
\begin{equation}\label{eq:kp-ot-map}
T(\boldsymbol{x})=\boldsymbol{x}-(\nabla h)^{-1}[\nabla \varphi (\boldsymbol{x})]
\end{equation}
\end{theorem}
This theorem
was first explained by Villani~\cite{villani2009optimal} and later organized by Lei et al. \cite{lei2019geometric,lei2020geometric}. Based on it, we further provide an explicit functional relationship between the OT map and Kantorovich's potential $\varphi$ for hyper-spheres case.

\begin{theorem}
\label{thm:our_optim_relation}
Suppose $\boldsymbol{X}\subset\mathbb{S}^{d}$, $\boldsymbol{Y}\subset\mathbb{S}^{d}$ are two subsets of $d$-dimensional hyper-sphere $\mathbb{S}^{d}$, and $\mu_S$ and $\nu_S$ are two probability measures defined on $\boldsymbol{X}$ and $\boldsymbol{Y}$, respectively. 
Given cost $c(\boldsymbol{x}, \boldsymbol{y}) = \log(2- \langle \boldsymbol{x},\boldsymbol{y}\rangle)$, there exists a Kantorovich's potential $\varphi$ such that the OT map $T:\boldsymbol{X}\to\boldsymbol{Y}$ can be represented as follows: 
\begin{equation}\label{eq:our-sot-map}
T(\boldsymbol{x})=\boldsymbol{x}-\frac{\nabla\varphi(\boldsymbol{x})}{1-\boldsymbol{x}^T\nabla\varphi (\boldsymbol{x})}
\end{equation}
\end{theorem}
\noindent\textbf{Proof.}  Assume $\boldsymbol{z}=\boldsymbol{x}-\boldsymbol{y}$, then, based on \textbf{Theorem~\ref{thm:kp_optim_relation}}, we get $h(\boldsymbol{z})=\log(2- \langle \boldsymbol{x},\boldsymbol{y}\rangle)$. The gradient and Hessian matrix of $h$ regarding $\boldsymbol{z}$ are as follows:
\begin{equation}\label{eq:gradient}
\nabla h(\boldsymbol{z})=\frac{\boldsymbol{x}-\boldsymbol{y}}{2-\boldsymbol{x}^T\boldsymbol{y}}
\end{equation}
\begin{equation}\label{eq:Hessian}
 h^{''}(\boldsymbol{z})=\frac{2(2-\boldsymbol{x}^T\boldsymbol{y})\boldsymbol{1}\boldsymbol{1}^T+\boldsymbol{z}\boldsymbol{z}^T}{(2-\boldsymbol{x}^T\boldsymbol{y})^2}
\end{equation}

The Hessian matrix in Eq.~\eqref{eq:Hessian} is positively definite, therefore $h$ is strictly convex. Suppose $(\boldsymbol{x}_0,\boldsymbol{y}_0)$ is a point in the support
of $T$, by definition $\varphi^c(\boldsymbol{y}_0)=\inf_{{\boldsymbol{x}}}(c(\boldsymbol{x}, \boldsymbol{y}_0)-\varphi(\boldsymbol{x}))$, hence, $\nabla_{\boldsymbol{x}} (c(\boldsymbol{x}, \boldsymbol{y}_0)-\varphi(\boldsymbol{x}))|_{\boldsymbol{x}=\boldsymbol{x}_0}=0$. Then, we get 
\begin{equation}\label{eq:grad_relat}
 \nabla\varphi(\boldsymbol{x}_0)=\nabla_{\boldsymbol{x}} c(\boldsymbol{x}_0, \boldsymbol{y}_0)=\nabla h(\boldsymbol{x}_0-\boldsymbol{y}_0)
\end{equation}
Combining Eq.~\eqref{eq:Hessian}, we obtain
\begin{align}
\label{eq:target_equation}
 \nabla\varphi(\boldsymbol{x})&=\frac{\boldsymbol{x}-\boldsymbol{y}}{2-\boldsymbol{x}^T\boldsymbol{y}}\\
 \boldsymbol{z}=[2-\boldsymbol{x}^T&(\boldsymbol{x}-\boldsymbol{z})] \nabla\varphi(\boldsymbol{x})=(1+\boldsymbol{x}^T\boldsymbol{z})\nabla\varphi(\boldsymbol{x})
\end{align}
Suppose $\boldsymbol{z}=\lambda\nabla\varphi(\boldsymbol{x})$, then we get
\begin{align*}
\label{eq:solve_equation}
 \lambda\nabla\varphi(\boldsymbol{x})&=(1+\lambda\boldsymbol{x}^T\nabla\varphi(\boldsymbol{x})) \nabla\varphi(\boldsymbol{x})\\
 \Rightarrow &
 \lambda(1-\boldsymbol{x}^T\nabla\varphi(\boldsymbol{x})) =1\\
 \Rightarrow &\lambda = \frac{1}{1-\boldsymbol{x}^T\nabla\varphi(\boldsymbol{x})}
 \Rightarrow\boldsymbol{z}=\frac{\nabla\varphi(\boldsymbol{x})}{1-\boldsymbol{x}^T\nabla\varphi(\boldsymbol{x})}\\
 \Rightarrow & \boldsymbol{y}=\boldsymbol{x}-\boldsymbol{z} = \boldsymbol{x}-\frac{\nabla\varphi(\boldsymbol{x})}{1-\boldsymbol{x}^T\nabla\varphi(\boldsymbol{x})}
\end{align*}

Theorem~\ref{thm:our_optim_relation} provides the explicit functional relationship between the SOT map and potential $\varphi$. 
However, the objective Eq.~\eqref{eq:dual_convex_form} is not amenable to standard stochastic optimization schemes due to the conjugate function $g^c$. To this end, we propose a novel minimax formulation in the following proposition, where we replace the $\boldsymbol{y}$ in $g^c$ with the transport map $T_f(\boldsymbol{x})$ defined in Eq.~\eqref{eq:SOTMap} on $\boldsymbol{X}$.

\begin{proposition}
\label{thm:our_optim_result}
Given two probability measures $\mu_S$ and $\nu_S$ which are defined on $\boldsymbol{X}\subset\mathbb{S}^{d}$ and $\boldsymbol{Y}\subset\mathbb{S}^{d}$, we have
\begin{equation}\label{eq:max-min-tx}
W_d = \sup_{\substack{g \in \mathtt{CVX}(\nu_S)}}\inf_{\substack{f \in \mathtt{CVX}(\mu_S)}} \mathbb{E}_{\nu_S}g(\boldsymbol{Y})+\mathcal{V}_{\mu_S}(f,g)
\end{equation}
where 
$\mathcal{V}_{\mu_S}(f,g)$ is a functional of $f,g$ defined as 
\begin{align*}
\mathcal{V}_{\mu_S}(f,g)= \mathbb{E}_{\mu_S}[\log (2-\langle \boldsymbol{X},T_f(\boldsymbol{X})\rangle)-g(T_f(\boldsymbol{X}))]
\end{align*}
where 
$T_f(\boldsymbol{X})$ is a functional of $f$ defined as
\begin{equation}
\label{eq:SOTMap}
T_f(\boldsymbol{x})=\boldsymbol{x}-\frac{\nabla f(\boldsymbol{x})}{1-\boldsymbol{x}^T\nabla f (\boldsymbol{x})}.
\end{equation}
\end{proposition}

Proposition~\ref{thm:our_optim_result} transforms the solution of the SOT map into the problem of optimizing the $W_d$ with $f$ and $g$ as variables. Its advantage lies in establishing a direct connection between the $f$ and $g$ and transforming the infimum calculation of c-transform into a parameter optimization process during the learning. Moreover, the gradient of potential $f$ induces the SOT map, which can fit discontinuous maps and allow sharp boundaries to be kept. 

According to Theorem~\ref{thm:our_optim_relation}, there exists a potential $f$ such that $T_f:\boldsymbol{X}\to\boldsymbol{Y}$ in Eq.~\eqref{eq:SOTMap} satisfies \eqref{eq:SOT}. This indicates that there exists an optimal pair $(f_0, g_0)$ achieving the infimum and supremum respectively, and $T_{f_0}$ is the OT map from $\mu_S$ to $\nu_S$. In other words, $T_{f_0}$ maps $\boldsymbol{X}$ to $\boldsymbol{Y}$ in a measure preserving manner, satisfying $T(\boldsymbol{X})=\boldsymbol{Y}$. Thus, the solution for Eq.~\eqref{eq:dual_convex_form} is also a solution for Eq.~\eqref{eq:max-min-tx}.  Based on this, the next pertinent question is how to construct a suitable $f$ and $g$ and effectively solve it in the text-to-3D generation task.

\begin{figure*}[htbp!]
\centering
\vspace{-3mm}
\subcaptionbox*{Input $\boldsymbol{X}$ Distribution\quad Target $\boldsymbol{Y}$ Distribution\qquad\qquad M \& V projection of SSW~\cite{bonet2022spherical}\qquad\qquad  M \& V projection of \textbf{Ours}\qquad}
{\includegraphics[width=0.99\linewidth]{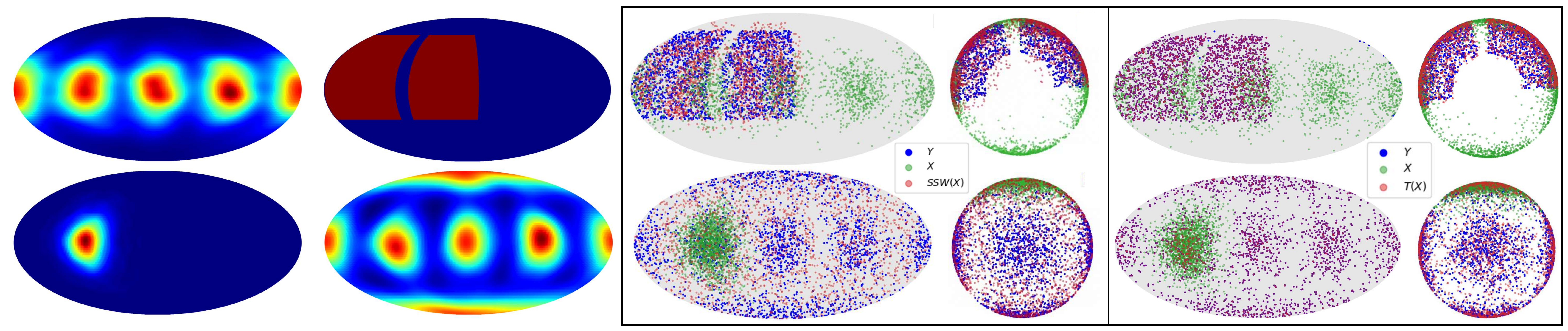}}
\caption{Visualization of two SOT map results. The first two columns indicate distributions of input and target (warmer color means higher probability). The last four columns are the visualization of the input point set $\boldsymbol{X}$, target point set $\boldsymbol{Y}$, and mapped results of SSW~\cite{bonet2022spherical} $SSW(\boldsymbol{X})$ and our SOT $T(\boldsymbol{X})$ by Mollweide~(M) projection and vertical~(V) projection. More overlap between \textcolor{pink}{pink SOT mapped results} and \textcolor{blue}{blue target} indicates \textbf{higher accuracy}. Here, the distribution of $T(\boldsymbol{X})$ is highly consistent with $\boldsymbol{Y}$ (overlap of pink and blue) and maintains clear boundaries.}
\label{fig:Ablation}
\vspace{-3mm}
\end{figure*}

\subsection{Text-to-3D generation with SOT map}
\label{sec:stylize}
Inspired by Makkuva et al.~\cite{makkuva2020optimal}, we instantiated the Kantorovich potential $f$ and $g$ with ICNN~\cite{amos2017input}, where ICNNs are a specific parametric class of convex functions with rich enough representation. The network architecture of potential $f$ and $g$ is shown in \cref{fig:network}. Specifically, in our experiment, we set $C_0=768$ and $C_1=1024$, respectively. Further, to improve the text-shape consistency, we add a fidelity loss to 
the objective~\eqref{eq:obj_f}, where the $d(T_f(\boldsymbol{x}),\boldsymbol{y})=cos^{-1}(T_f(\boldsymbol{x}),\boldsymbol{y})$ is the spherical distance between the output and the paired image feature. $\lambda$ is the weight, which was set to 1 in our experiment. Finally, the desired $f$ and $g$ pair is obtained by a training procedure with alternate optimization: \textit{i) Potential $f$ is optimized by Eq. \eqref{eq:obj_f};} \textit{ii) Potential $g$ is optimized by Eq. \eqref{eq:obj_g}. } 
\begin{align}
\inf_{f \in \mathtt{CVX}(\mu_S)}[\mathcal{V}_{\mu_S}(f,g)+\lambda\mathbb{E}_{\mu_S}d(T_f(\boldsymbol{X}),\boldsymbol{Y})],\label{eq:obj_f}\\
\sup_{g \in \mathtt{CVX}(\nu_S)} [\mathbb{E}_{\nu_S}g(\boldsymbol{Y})-\mathbb{E}_{\mu_S}g(T_f(\boldsymbol{X}))].
\label{eq:obj_g}
\end{align}

After getting the well-optimized $f$, the SOT map $T_f(\boldsymbol{x})$ from text feature hyper-sphere $\boldsymbol{X}\subset\mathbb{S}^{767}$ to image feature hyper-sphere $\boldsymbol{Y}\subset\mathbb{S}^{767}$, can be obtained with Eq.~\eqref{eq:SOTMap} to achieve semantic alignment. 
For input texts, we utilize CLIP to obtain the initial embedded text features. In previous approaches, text features were directly used as image features for generation. 
To bridge the gap between the image and text features, we apply the obtained SOT map to project text feature ${\boldsymbol{X}}$ to the aligned $\hat{\boldsymbol{Y}}$ in the CLIP embedding space.
Apart from spherical distance, the SOT map also aligns the two modalities from the view of global distribution. Thereafter, as illustrated in \cref{fig:framework}, a conditional generator is followed to transform the output feature to signed distance functions (SDFs) and texture color. To construct a triangle mesh, a differentiable Marching Cubes~\cite{shen2021deep} is adopted.


\section{Experiments}\label{sec:experiments}
We conduct comprehensive experiments to evaluate the performance of our model. First, we demonstrate the effectiveness of our SOT module, which forms the core of HOTS3D, as described in~\cref{sec:Visual-SOT}. Next, we showcase HOTS3D's capability in text-to-3D generation using the Text2Shape~\cite{chen2018text2shape} and Objaverse~\cite{deitke2023objaverse} datasets, as detailed in~\cref{sec:Text2Shape}. Finally, we perform ablation studies to analyze the impact of our SOT module on 3D generation in ~\cref{sec:Ablation}.

\subsection{Effectiveness of the SOT Map} \label{sec:Visual-SOT}
To demonstrate the effectiveness of the proposed SOT algorithm in section~\ref{sec:SOTmap}, we conducted comparative experiments with SSW~\cite{bonet2022spherical} on synthesized spherical data, and the visualization results are shown in \cref{fig:Ablation}. 

Here, we employed Mollweide projection~\cite{lapaine2011mollweide} and vertical projection to visualize the results. The Mollweide projection is an area-preserving~\cite{zhao2013area} projection generally used for maps of the world or celestial sphere. It trades the accuracy of angle and shape for the accuracy of proportions in area, making it useful for maps depicting global distributions. Specifically, in the first three and fifth columns of \cref{fig:Ablation}, we used Mollweide projection, and in the fourth and last columns, we employed vertical projection to provide different viewing perspectives. In the first row of \cref{fig:Ablation}, the source domain data $\boldsymbol{X}$ follows four modes \textit{Von Mises-Fisher} (\textit{vMF})~\cite{mardia1975statistics} distribution, while the target data $\boldsymbol{Y}$ follows \textit{uniform} distribution on two discontinuous one-eighth spheres. In the second row, the source $\boldsymbol{X}$ follows single mode \textit{vMF} and the target  $\boldsymbol{Y}$ follows six modes \textit{vMF}. 

The \textit{vMF} distribution, often regarded as the spherical counterpart to the Gaussian distribution, is a distribution on $S^{d-1}$ characterized by a concentration parameter $\kappa>0$ and a location parameter $\mu\in S^{d-1}$ through the density  
\begin{equation*}
    \forall \theta\in S^{d-1},\ f_{\mathrm{vMF}}(\theta;\mu,\kappa) = \frac{\kappa^{d/2-1}}{(2\pi)^{d/2} I_{d/2-1}(\kappa)} \exp(\kappa \mu^T\theta).
\end{equation*}
Where $I_\nu(\kappa)=\frac{1}{2\pi}\int_0^\pi \exp(\kappa\cos(\theta))\cos(\nu\theta)\mathrm{d}\theta$ is the modified Bessel function. Specifically, we set the mean direction ($\mu$ in Euclidean space) to $(1,0,0)$, $(0,1,0)$, $(-1,0,0)$ and $(0,-1,0)$ respectively, and then adopt commonly used open source code\footnote{\url{ https://github.com/dlwhittenbury/von-Mises-Fisher-Sampling/}} to generate a mixed vMF distribution with 4 modes. For the target data $Y$, we uniformly sampled on two discontinuous one-eighth spheres, whose elevation $\theta$ and azimuth $\varphi$ ranges are $\frac{\pi}{3}\leq\theta\leq\frac{5\pi}{6},0\leq\varphi\leq\frac{\pi}{2}$ and $\frac{\pi}{3}\leq\theta\leq\frac{5\pi}{6},\frac{\pi}{2}+\frac{\pi}{12}\leq\varphi\leq\pi+\frac{\pi}{12}$. For another experiment, we set the mean direction to $(1,0,0)$, $(0,1,0)$, $(0,0,1)$, $(-1,0,0)$, $(0,-1,0)$, and $(0,0,-1)$ to generate a mixed vMF distribution with 6 modes, and take it as the target distribution. Then, we set the mean direction as $(1,0,0)$ to generate a single mode vMF and take it as the source distribution.

In this experiment,  we instantiated the $f$ and $g$ in~\cref{fig:network} with a 4-layer ICNN~\cite{amos2017input}, including 2 \textit{Convex Block}. We use Adam~\cite{kingma2014adam} with learning rate in $10^{-4}$ to optimize the objective of \eqref{eq:obj_f} and \eqref{eq:obj_g}. Here, $C_0=3$, $C_1=64$, $\lambda=0$.

The results show that, compared to SSW, our SOT map not only aligns the target distribution more accurately but also preserves the boundaries more effectively. This demonstrates the superiority of our method in handling spherical distribution alignment, particularly in the presence of discontinuities in the data distribution.

\subsection{Text-to-3d Generation}\label{sec:Text2Shape}
\subsubsection{Setup}\label{sec:Text2Shapesetup}
\noindent\textbf{Datasets.} To benchmark HOTS3D against prior methods, we use the widely adopted ShapeNet~\cite{chang2015shapenet} and Objaverse~\cite{deitke2023objaverse} datasets. For ShapeNet, we follow the setup in baselines~\cite{cheng2023sdfusion,li2023diffusion,Sanghi_2022_CVPR} and choose two specific categories `chair' and `table', and the corresponding text prompts are from dataset Text2Shape~\cite{chen2018text2shape}. 
The ground truth images, including those used for training and evaluation, are generated by selecting the corresponding meshes from ShapeNet and rendering 20 views for each shape. We employ Blender~\cite{blender} to render with a simple lighting and material setup: all models were rendered with a fixed lighting configuration that supports only diffuse and ambient shading. We use the first 1800 text prompts of the Text2Shape test dataset for testing.

{For Objaverse, we use the paired image-text dataset from \cite{luo2024scalable} to train and test our model. Specifically, they employ Blender~\cite{blender} to render 20 images for each 3D mesh and BLIP2~\cite{li2023blip} for captioning. All image sizes are uniformly set to $512\times512$. The text prompts we utilized for testing are 2000 texts selected at equal intervals from the entire dataset.}

\textbf{Implementation Details.} For HOTS3D, we only need to train the SOT module, while the other components remain frozen. Specifically, we instantiate $f$ and $g$ with a 10-layer ICNN, which includes 8 \textit{Convex Blocks} (see~\cref{fig:network}), and train the SOT map by extracting text embeddings and corresponding image embeddings from shape renderings using the CLIP ViT-L/14 encoder~\cite{radford2021learning}. The conditional generator is based on an open-source Shap$\cdot$E implementation~\cite{jun2023shap}.

More specifically, the generator is implemented as the composite of a conditional diffusion model and a decoder. The conditional diffusion model is with a \textit{MLP+Transformer+ MLP} architecture: the first \textit{MLP} integrates the conditional CLIP embedding, and the second \textit{MLP} processes the implicit field. The NeRF-based decoder is then used to render the implicit field and extract the SDF for mesh generation. The potential $f$ and $g$ are optimized using the objectives defined in \eqref{eq:obj_f} and \eqref{eq:obj_g}, with the weight $\lambda$ empirically set to 1. We use a learning rate of $10^{-4}$ and the Adam optimizer~\cite{kingma2014adam}. Training the SOT map on the Text2Shape dataset~\cite{chen2018text2shape} takes approximately 28 hours on a Tesla A100 GPU.

\textbf{Baseline Methods.} {We compare HOTS3D with thirteen baseline approaches, including end-to-end models: SDFusion~\cite{cheng2023sdfusion}, Shap·E~\cite{jun2023shap}, TAPS3D~\cite{wei2023taps3d}, EXIM~\cite{liu2023exim}, XCube~\cite{ren2024xcube}, Michelangelo~\cite{zhao2024michelangelo},  Ln3diff~\cite{lan2024ln3diff}, Kiss3DGen~\cite{lin2025kiss3dgen}, and Hunyuan3D~\cite{yang2024hunyuan3d}, and optimization-based methods: 3DFuse~\cite{seolet}, PCLIPNeRF~\cite{lee2022understanding}, ProlificDreamer~\cite{wang2023prolificdreamer}, RichDreamer~\cite{qiu2023richdreamer}. For fair comparisons, we use the default settings, official implementations, and pretrained models for all baseline methods.} 

\textbf{Evaluation Metrics.} For a comprehensive quantitative evaluation, we employ FID~\cite{heusel2017gans}, P-FID~\cite{nichol2022point}, LFD~\cite{chen2003visual} score, Chamfer Distance (CD), {and F-Score to assess the quality of the generated shapes. FID measures the distance between the rendered image distribution and ground truth distribution. LFD quantifies the dissimilarity between a ground truth mesh and a generated mesh by comparing rendered images from 60 viewpoints, while CD measures the distance between point clouds sampled from the two shapes. P-FID is a modified version of FID tailored for point clouds, designed to measure the distance between the distribution of generated point clouds and that of real point clouds. F-Score evaluates the similarity between the generated and the ground-truth surface.}
Moreover, to measure the consistency between the generated shapes and input text, we report the CLIP R-Precision~\cite{park2021benchmark}. This metric quantifies the percentage of generated images the CLIP encoder associates with the correct text prompt used for generation. We followed the original settings of CLIP R-Precision~\cite{park2021benchmark} to calculate the top-10 retrieval accuracy when retrieving the matching text from 100 candidates using the generated image as a query. For 3D metrics P-FID, CD, and F-Score, we uniformly sample 4,096 points from meshes.

\begin{table}[t]
\footnotesize
\caption{Quantitative comparison on Text2Shape. Two distinct models are utilized to compute the CLIP R-Precision. S-A100 denotes the average time (in seconds) it takes to process a text on the NVIDIA Tesla A100 GPU. The methods of the last two rows are optimization-based.}
\vspace{-2mm}
\renewcommand{\arraystretch}{1.2}
\renewcommand{\tabcolsep}{1.2mm}
\centering
\begin{tabular}{@{}lccccc@{}}
\toprule
\multirow{2}{*}{Method}  & \multicolumn{2}{c}{CLIP R-Precision $\uparrow$}& \multirow{2}{*}{FID $\downarrow$} & \multirow{2}{*}{LFD $\downarrow$} & \multirow{2}{*}{S-A100$\downarrow$}\\
                   & ViT-B/32       & ViT-L/14  & \\ \hline
Shap$\cdot$E        & 0.698  &0.683  &92.1& 6945 & \textbf{7.1}  \\
SDFusion     &    0.723    &      0.728&173.8& 5929 &  8.9 \\
TAPS3D   &  0.666  &   0.676 &79.2 & 6530 & 9.7 \\
EXIM& 0.714  & 0.721   &45.3 & 6005 & 21.7 \\
Michelangelo& 0.746  &0.752    & 100.9& 6632 &7.8  \\
Ours     &     \textbf{0.793} & \textbf{0.801} & \textbf{42.6}& \textbf{5349} &10.2   \\    
\hline
PCLIPNeRF    &  \textbf{0.828}  & 0.810   &172.3  & -&  1597.9\\
3DFuse    &  0.823  & \textbf{0.829}   &148  & -&  2417.6\\
\bottomrule
\end{tabular}
\label{tab:text23D}
\end{table}

\begin{figure}[htb!]
\centering
\vspace{-3mm}
\subcaptionbox*{Kiss3DGen\ \  Ln3Diff\ \   TAPS3D\ \  SDFusion \quad Shap·E \quad \textbf{Ours}\ }
{\includegraphics[width = 1.0\linewidth]{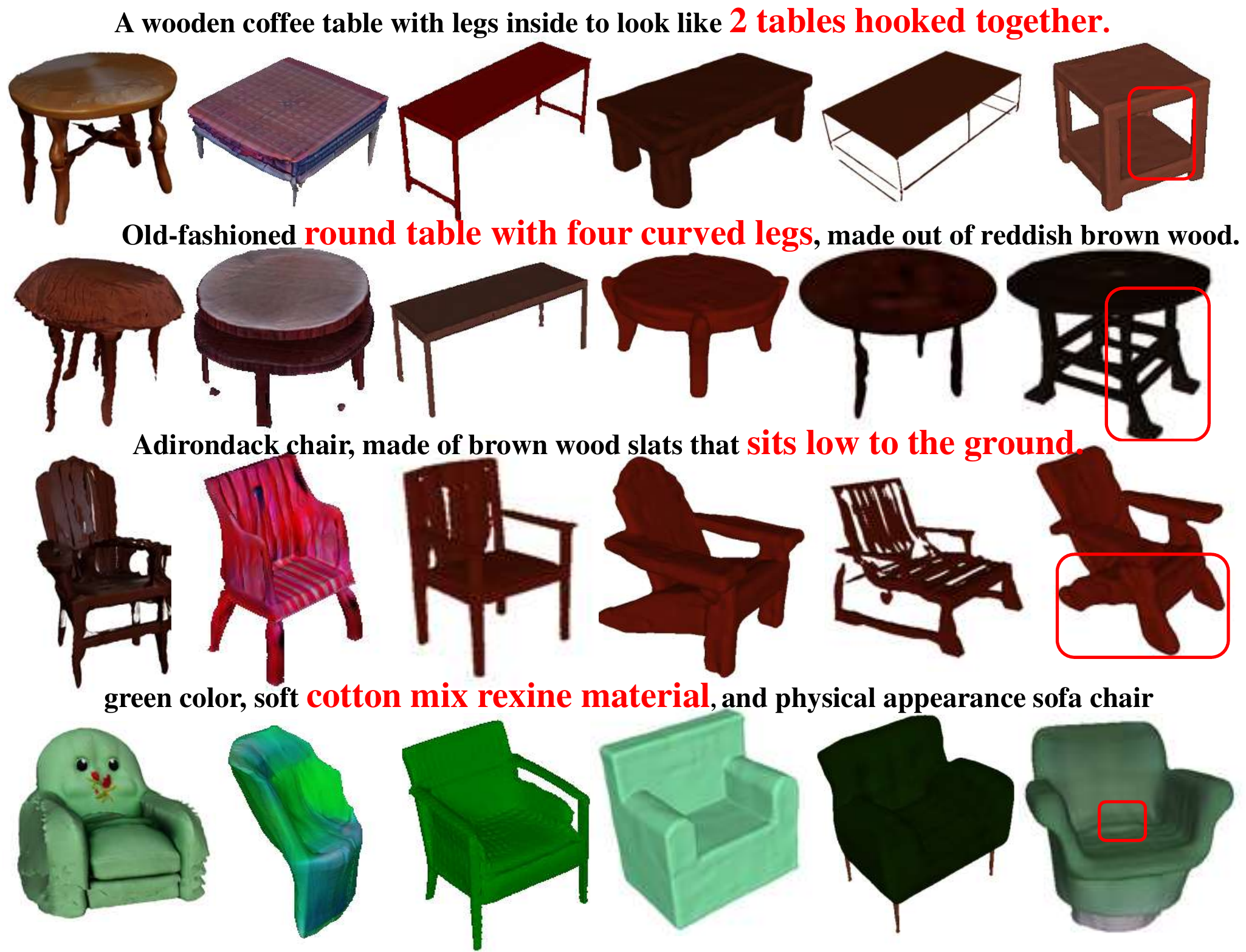}}
\vspace{-3mm}
\caption{{Qualitative comparison on Text2Shape. Red boxes on results match the distinctive descriptions highlighted in red.}}
\label{fig:text23D}
\vspace{-5mm}
\end{figure}
\subsubsection{Quantitative Results} 
{For quantitative comparisons, we report the CLIP retrieval precision and other shape assessment metrics in \cref{tab:text23D} and \cref{tab:Objaverse}, and both the CLIP ViT-B/16 and CLIP ViT-B/32 encoders are employed as retrieval models. Compared to end-to-end models, our HOTS3D achieves the highest CLIP R Precision with both retrieval models. Compared to optimization based methods, our HOTS3D can achieve comparable performance with significantly lower time costs. By leveraging the SOT module, our method bridges the modality gap and thus can generate more faithful 3D assets to the input text. Moreover, by aligning text features to multiple image features from the same object, we can compensate for the missing detailed information in texts composed of high-level semantics. Consequently, our method exhibits better scores than these baseline methods.}

To investigate whether semantic alignment can benefit shape quality, we evaluated the dissimilarity between real meshes and generated meshes using the LFD metric. Since LFD is a mesh-based metric and the final outputs of PCLIPNeRF and 3DFuse are rendered images from multiple views, they cannot be used to compute LFD scores. As seen in ~\cref{tab:text23D}, our method achieved the best LFD score, indicating that it produces high-quality 3D shapes consistent with the target objects. We also present a comparison of the generation efficiency of a single text. As shown in the \cref{tab:text23D}, optimization-based methods~\cite{lee2022understanding,seolet} require training the model for each input text, so it often takes tens of minutes to process a single text well. In comparison, our method can generate 3D shapes that are high-quality and highly consistent with input texts in just about 10 seconds. Compared to Shap·E, the SOT map of our method only incurs an additional expense of 3.2 seconds, which is marginal for 3D generation, but improves substantially in all metrics, especially in FID score. Compared with the SDS-based optimization method 3DFuse~\cite{seolet}. Our HOTS3D demonstrates comparable scores while significantly reducing time costs.

Moreover, to evaluate the performance over diverse high-quality 3D datasets, we also include the experiments on Objaverse and present the quantitative comparisons in Tab.~\ref{tab:Objaverse}. Our method achieved the highest CLIP-R score, indicating superior semantic consistency with the input text compared to baselines. The key difference between our approach and Shap$\cdot$E lies in the incorporation of the proposed SOT module. By leveraging the SOT map, our method effectively bridges the modality gap, enabling the generation of 3D structures that are more faithful to the input text. Additionally, by aligning text features with multiple image feature regions of the same object, our approach compensates for the lack of fine-grained details often missing in high-level semantic text descriptions. As a result, our method outperforms others on 3D metrics such as LFD, P-FID, CD, and F-Score, indicating lower reconstruction error and improved overall shape quality. These results suggest that enhancing semantic consistency also leads to improved shape fidelity.

\begin{table}[t]
\caption{Quantitative comparison on Objaverse.}
\vspace{-2mm}
\footnotesize
\renewcommand{\arraystretch}{1.2}
\renewcommand{\tabcolsep}{0.98mm}
\centering
\begin{tabular}{@{}lcccccc@{}}
\toprule
\multirow{2}{*}{Method}  & \multicolumn{2}{c}{CLIP R-Precision $\uparrow$}& \multirow{2}{*}{P-FID $\downarrow$} & \multirow{2}{*}{LFD $\downarrow$} & \multirow{2}{*}{CD$\downarrow$}&\multirow{2}{*}{F-Score$\uparrow$} \\
                   & ViT-B/32       & ViT-L/14  & \\ \hline
Shap$\cdot$E  & 0.824  & 0.845 &3.0 &7067 &  0.159 &  0.909 \\
SDFusion     & 0.712 & 0.725 & 6.8 & 9447    &0.128 & 0.747 \\
TAPS3D    & 0.807   &0.803  & 20.2 &10644 & 0.119 & 0.869  \\
Xcube & 0.790  &0.796 & 3.0  &7837 & 0.150 & 0.957  \\
Michelangelo& 0.864  &  0.883  &3.2 &10321  &0.154 & 0.893 \\
Ln3diff   & 0.835   & 0.862 & 2.8 & 7516& 0.139 & 0.928  \\
Hunyuan3D & 0.878  &0.901 & 5.7 &7671 & 0.150 &  0.915 \\
Kiss3DGen&  0.800 & 0.846   &7.1 & 8203 & 0.142& 0.953 \\
Ours     &\textbf{0.900}     & \textbf{0.911} &\textbf{2.5} & \textbf{4640} & \textbf{0.068}  & \textbf{0.976}\\    
\bottomrule
\end{tabular}
\label{tab:Objaverse}
\vspace{-2mm}
\end{table}

\subsubsection{Qualitative Results}
The qualitative comparison on Text2Shape is presented in \cref{fig:text23D}, indicating that baselines~\cite{Sanghi_2022_CVPR,jun2023shap, cheng2023sdfusion, wei2023taps3d} encounter challenges in generating precise and realistic 3D objects, resulting in semantic inconsistency or incomplete 3D shapes. {The results of Kiss3DGen still have significant room for improvement in geometry and shape. Ln3Diff can produce 3D content with richer textures and colors, but struggles with semantic consistency.} TAPS3D often produces simple tables and chairs, and Shap·E lacks geometric control, both needing better semantic alignment. Although SDFusion performs more reasonably, it still shows text-shape mismatches. Our method, integrating the SOT map, produces more coherent and higher-quality 3D objects, matching the text closely, as seen in examples like ``2\textit{ tables hooked together}''.

\begin{figure*}[htb!]
\vspace{-3mm}
\centering
    \includegraphics[
    height = 0.65\linewidth,
  width = .9\linewidth]{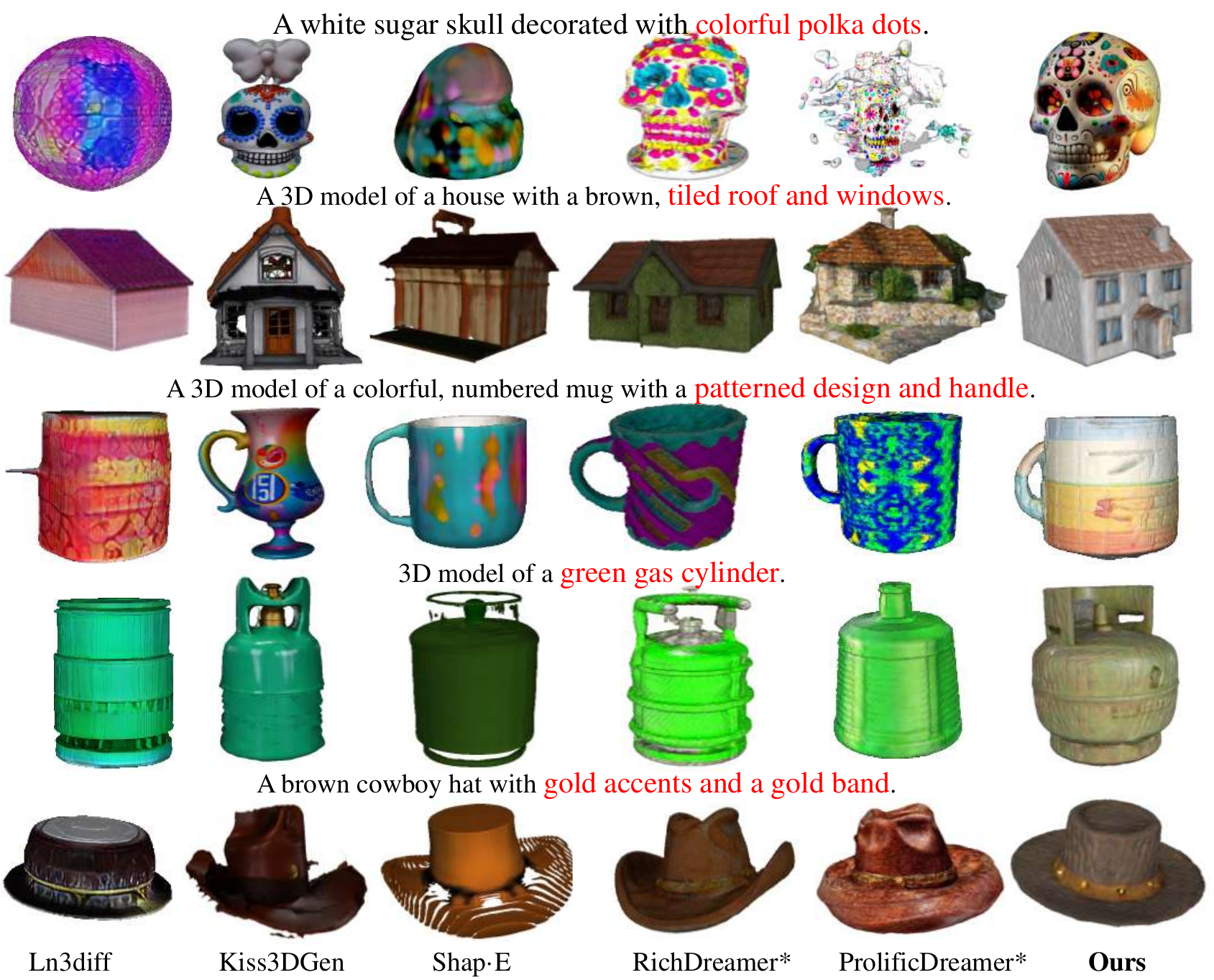}
\vspace{-4mm}
\caption{{Qualitative comparison on Objaverse. Red boxes on the results match the distinctive descriptions highlighted in red.}}
\label{fig:objaverse}
\vspace{-3mm}
\end{figure*}

The qualitative comparison in Fig.~\ref{fig:objaverse} highlights results on Objaverse. {Ln3diff produces outputs that deviate significantly from the input text. Kiss3DGen and Shap·E generate decent geometry and texture but struggle with semantic consistency. ProlificDreamer$^\ast$ and RichDreamer$^\ast$, as optimization-based methods, achieve better semantic alignment but require several hours per prompt. In contrast, our method generates 3D models that align closely with the input text in an end-to-end fashion, while preserving higher quality and detail, demonstrating that semantic alignment can enhance the geometric quality of generated models.}

\subsubsection{User studies}
\begin{figure}
\centering
\includegraphics[width=1.0\linewidth]{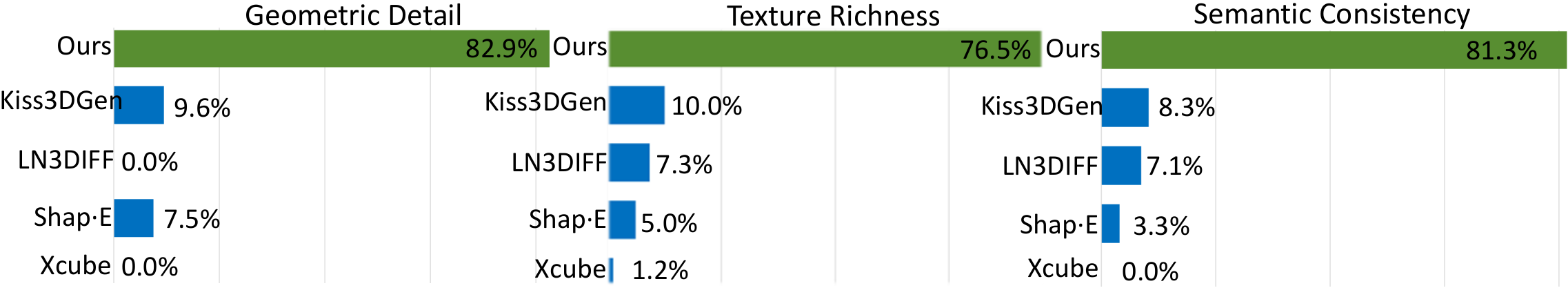}
\caption{{User studies comparing HOTS3D with state-of-the-art methods show a strong preference for HOTS3D in terms of geometry quality, texture quality, and semantic consistency.}}
\label{fig:userstudyRes}
\vspace{-3mm}
\end{figure}

To enable a thorough comparison, we conducted a comprehensive user study focused on three key aspects: geometry quality, texture quality, and alignment with the text prompt. The study used a test set of 12 text prompts and involved 80 participants.  As illustrated in Fig.~\ref{fig:userstudyRes}, HOTS3D outperforms other methods in appearance, geometry, and semantic consistency for text-to-3D generation. Specifically, HOTS3D received 82.9\% of the votes for geometry, 91.6\% for texture, and 81.3\% for semantic consistency. Considering the cost of participant involvement, we conducted the comparative experiments on a limited set of results (12 text prompts). Nonetheless, the observed user preferences still provide clear evidence of HOTS3D’s effectiveness.

\subsubsection{Ablation Studies}\label{sec:Ablation}

The proposed SOT Map can be seamlessly integrated into existing text-to-3D methods, enhancing semantic consistency in a plug-and-play manner. To assess the effectiveness of the SOT module in improving semantic alignment, we incorporated it into three baseline models: Shap$\cdot$E~\cite{jun2023shap}, Xcube~\cite{ren2024xcube}, and Ln3diff~\cite{lan2024ln3diff}. We then conducted a quantitative comparison of their results, as shown in Tab.~\ref{tab:ablationSOT}. The metrics show that adding the SOT module improves CLIP R-Precision, indicating enhanced semantic-text consistency. Furthermore, improvements in FID, CD, and F-Score highlight the module’s effectiveness in shape reconstruction.

\begin{table}[t]
\footnotesize
\caption{Ablation Study with the SOT Map: We incorporate the SOT module into Shap$\cdot$E, Xcube, and Ln3diff, compare metrics before and after integration on Objaverse.}
\vspace{-2mm}
\renewcommand{\arraystretch}{1.2}
\renewcommand{\tabcolsep}{1.2mm}
\centering
\begin{tabular}{@{}lccccc@{}}
\toprule
\multirow{2}{*}{Method}  & \multicolumn{2}{c}{CLIP R-Precision $\uparrow$}& \multirow{2}{*}{P-FID $\downarrow$} & \multirow{2}{*}{CD $\downarrow$} & \multirow{2}{*}{F-Score$\uparrow$}\\
                   & ViT-B/32       & ViT-L/14  & \\ \hline
Xcube     & 0.790&  0.796 & 3.0 & 0.150  & 0.953  \\
Xcube+SOT   &  0.873 &  0.909   & 2.6 &  0.150 & 0.960  \\
\hline
Ln3diff & 0.835  &  0.862  & 2.8& 0.139 & 0.954 \\
Ln3diff+SOT & 0.863  & 0.907   & 4.4& 0.112 & 0.968 \\ 
\hline
Shap$\cdot$E    & 0.824  & 0.845 &3.0 &  0.159 & 0.820   \\
Shap$\cdot$E+SOT & 0.900 & 0.911 & 2.5&  0.068 & 0.955  \\
\bottomrule
\end{tabular}
\label{tab:ablationSOT}
\vspace{-3mm}
\end{table}

\section{Conclusion}
\label{sec:conclusion}
In this paper, we propose a novel text-to-3D generation framework in combination with hyper-spherical OT, \ie \textbf{HOTS3D}, which can generate diverse and plausible 3D shapes from text prompts. Our method constructs the SOT map from the text feature hyper-sphere to the image feature hyper-sphere, achieving alignment between text and image features, while compensating to some extent for the lack of detailed information in high-concise text prompts. Then, a pre-trained diffusion-based conditional generator was utilized to transform the aligned feature into 3D content. 
Experimental results on synthetic data confirm the effectiveness of the proposed SOT algorithm for distribution alignment. Quantitative and qualitative comparisons on Text2Shape and Objaverse datasets demonstrate that our method can generate high-quality 3D shapes that accurately reflect text semantics. End-to-end distribution alignment allows the SOT module to be seamlessly integrated into other models, improving semantic consistency, while also making HOTS3D highly efficient with strong generalization and diversity. 

One limitation of our approach is that it frequently generates inconsistent outputs when handling long texts. Another issue is that optimizing two convex functions simultaneously during training can lead to unnecessary computational overhead. In future work, we plan to simplify the computational architecture and explore more powerful vision language models to improve output consistency.

\bibliographystyle{IEEEtran}
\bibliography{reference}

\vspace{-3mm}
\begin{IEEEbiography}[{\includegraphics[width=0.90in,height=1.20in,clip,keepaspectratio]{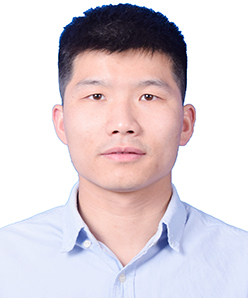}}]{Zezeng Li}
	received a B.S. degree from Beijing University of Technology (BJUT) in 2015 and a Ph.D. degree from Dalian University of Technology (DUT) in 2024. His research interests include image processing, point cloud processing, and generative model.
\end{IEEEbiography}

\begin{IEEEbiography}[{\includegraphics[width=1in,height=1.25in,clip,keepaspectratio]{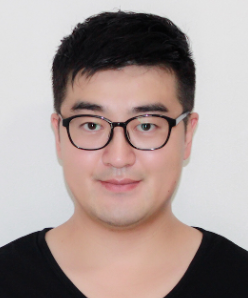}}]{Weimin Wang}
received a B.S. degree from Shanghai Jiao Tong University in 2009 and a Ph.D. degree from Nagoya University in 2017. He is currently an Associate Professor at Dalian University of Technology. He researches machine perception, scene understanding and interaction with different scales of the 3D physical world, aiming to enable machines to see the world in super vision.
\end{IEEEbiography}

\begin{IEEEbiography}[{\includegraphics[width=1in,height=1.25in,clip,keepaspectratio]{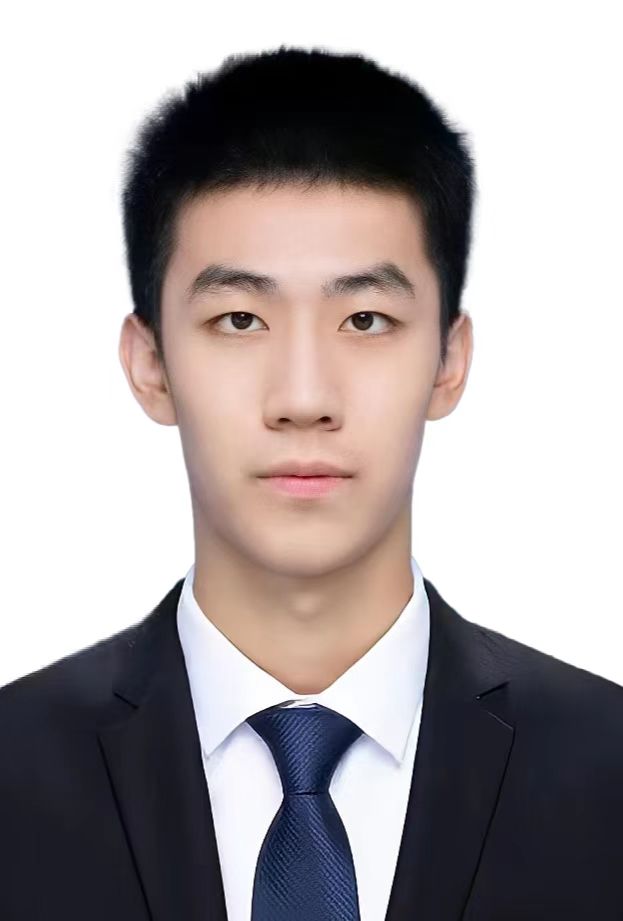}}]{Yuming Zhao} is a senior undergraduate student at Dalian University of Technology, with research interests in computer vision and mesh generation.
\end{IEEEbiography}

\begin{IEEEbiography}[{\includegraphics[width=1in,height=1.25in,clip,keepaspectratio]{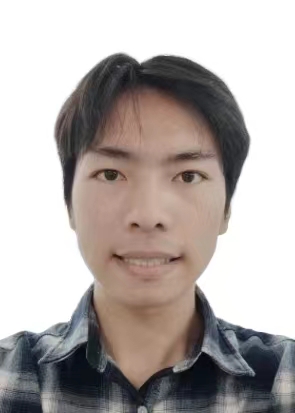}}]{Wenhai Li}
	received a B.S. degree from Nanchang HangKong University in 2022. He's a third-year master's student at Dalian University of Technology. His research interests include computer vision and mesh generation.
\end{IEEEbiography}

\begin{IEEEbiography}[{\includegraphics[width=1in,height=1.25in,clip,keepaspectratio]{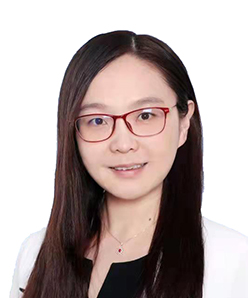}}]{Na Lei}
	received her B.S. degree in 1998 and a Ph.D. degree in 2002 from Jilin University. Currently, she is a professor at Dalian University of Technology. Her research interest is the application of modern differential geometry and algebraic geometry to solve problems in engineering and medical fields. She mainly focuses on computational conformal geometry, computer mathematics, and its applications in computer vision and geometric modeling.
\end{IEEEbiography}

\begin{IEEEbiography}[{\includegraphics[width=1in,height=1.25in,clip,keepaspectratio]{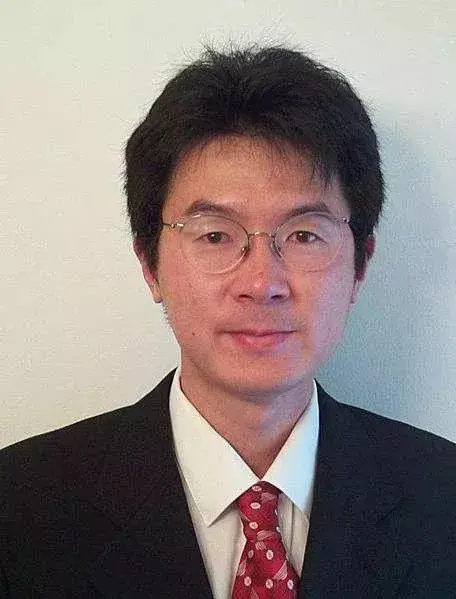}}]{Xianfeng Gu}
    received a B.S. degree from Tsinghua University and a Ph.D. degree from Harvard University. He is now a tenured professor in the Department of Computer Science and Applied Mathematics at the State University of New York at Stony Brook. Professor Gu's team combines differential geometry, algebraic topology, Riemann surface theory, partial differential equations, and computer science to create a cross-disciplinary "computational conformal geometry", which is widely utilized in computer graphics, computer vision, 3D geometric modeling and visualization, wireless sensor networks, medical images, and other fields.
\end{IEEEbiography}

\end{document}